\documentclass[sigconf]{aamas} 

\usepackage{balance} 

\setcopyright{ifaamas}
\acmConference[AAMAS '26]{This work as been accepted as an Extended Abstract in the Proc.\@ of the 25th International Conference
on Autonomous Agents and Multiagent Systems (AAMAS 2026)}{May 25 -- 29, 2026}
{Paphos, Cyprus}{C.~Amato, L.~Dennis, V.~Mascardi, J.~Thangarajah (eds.)}
\copyrightyear{2026}
\acmYear{2026}
\acmDOI{}
\acmPrice{}
\acmISBN{}

\acmSubmissionID{627}

\title[AAMAS-2026 Formatting Instructions]{Deep Meta Coordination Graphs for \\Multi-agent Reinforcement Learning} 

\author{Nikunj Gupta}
\affiliation{
  \institution{University of Southern California}
  \country{Los Angeles, CA, USA}}
\email{nikunj@usc.edu} 

\author{James Zachary Hare}
\affiliation{
    \institution{DEVCOM Army Research Laboratory}
    \country{Adelphi, MD, USA}}
\email{james.z.hare.civ@army.mil} 

\author{Jesse Milzman}
\affiliation{
    \institution{DEVCOM Army Research Laboratory}
    \country{Adelphi, MD, USA}}
\email{jesse.m.milzman.civ@army.mil} 

\author{Rajgopal Kannan}
\affiliation{
    \institution{DEVCOM ARL Army Research Office}
    \country{Los Angeles, CA, USA}}
\email{rajgopal.kannan.civ@army.mil} 

\author{Viktor Prasanna}
\affiliation{
  \institution{University of Southern California}
  \country{Los Angeles, CA, USA}}
\email{prasanna@usc.edu}

\begin{abstract}

This paper presents \textit{\algosmall} (\algoabb) for learning cooperative policies in multi-agent reinforcement learning (MARL). Coordination graph formulations encode local interactions and accordingly factorize the joint value function of all agents to improve efficiency in MARL. Through \algoabb, we dynamically compose what we refer to as \textit{meta coordination graphs}, to learn a more expressive representation of agent interactions and use them to integrate agent information through graph convolutional networks. The goal is to enable an evolving coordination graph to guide effective coordination in cooperative MARL tasks. The graphs are jointly optimized with agents' value functions to learn to implicitly reason about joint actions, facilitating the end-to-end learning of interaction representations and coordinated policies. We demonstrate that \algoabb\ consistently achieves state-of-the-art coordination performance and sample efficiency on challenging cooperative tasks, outperforming several prior graph-based and non-graph-based MARL baselines. Through several ablations, we also isolate the impact of individual components in \algoabb, showing that the observed improvements are due to the meaningful design choices in this approach. We also include an analysis of its computational complexity to discuss its practicality in real-world applications. All codes can be found here: {\color{blue}\url{https://github.com/Nikunj-Gupta/dmcg-marl}}. 

\end{abstract}

\keywords{Multi-agent deep reinforcement learning, coordination, graph representations} 

\newcommand{\BibTeX}{\rm B\kern-.05em{\sc i\kern-.025em b}\kern-.08em\TeX}
\usepackage{multirow}
\usepackage{enumitem}
\usepackage{booktabs} 
\usepackage{array}    
\usepackage{amsmath}
\usepackage{xcolor}         
\usepackage{amsfonts}
\usepackage{subfig}
\usepackage{bm}
\usepackage{tikz}
\usepackage{cleveref}
\usetikzlibrary{positioning, shapes.geometric, calc, decorations.pathreplacing}

\newcommand{\context}[1]{}

\newcommand{\edits}[1]{#1} 
\newcommand{\newedits}[1]{#1} 
 
\newcommand{\algoabb}{\textsc{DMCG}} 
\newcommand{\algosmall}{\textit{deep meta coordination graphs}} 
\newcommand{\algotitle}{Deep meta coordination graphs}

\begin{document}


\pagestyle{fancy}
\fancyhead{}


\maketitle 


\section{Introduction} 
\label{sec:introduction}

Multi-agent reinforcement learning (MARL) has become central to many real-world applications, such as warehouse robots \cite{orr2023multi,keith2024review,gupta2025hammer}, drone swarms \cite{alharbi2024reinforcement}, autonomous vehicles \cite{zhang2024multi}, and many other domains \cite{ning2024survey,li2022applications,afsar2022reinforcement}, where multiple agents must cooperate under partial observability and uncertainty. Value-decomposition methods~\cite{sunehag2018value,rashid2018qmix,rashid2020weighted,wang2021qplex,son2019qtran} have advanced MARL by scaling training through centralized training with decentralized execution \cite{foerster2018counterfactual,wen2019probabilistic}, avoiding the exponential joint action space of fully centralized learning \cite{oroojlooy2023review} and mitigating the non-stationarity of naive decentralization \cite{tan1993multi}. However, such fully factorized value functions have been shown to struggle with credit assignment and are prone to relative overgeneralization \cite{panait2006biasing}, where the same individual action can appear uninformative because its payoff depends on whether other agents coordinate. For example, in the Pursuit task \cite{benda1986optimal,stone2000multiagent,son2019qtran}, a prey can only be caught if at least two predators choose the catch action at the same time. A single predator catching gives no reward, but two acting together succeed. Fully factorized methods average over such contexts and fail to capture this dependency. Similar miscoordination also occurs in real-world settings, such as when robots must lift an object together or autonomous cars must decide whether to merge at an intersection. 

The approach of coordination graphs (CGs) \cite{guestrin2002coordinated,kok2006collaborative,bohmer2020deep,li2021deep,kang2022non,kortvelesy2022qgnn,duan2024group,phan2021vast} represents such localized interactions and decomposes the global value into utilities and pairwise (or small-set) payoff terms, enabling agents to reason about how subsets of teammates affect their outcomes. Methods such as deep CGs (DCG) \cite{bohmer2020deep} model fixed pairwise dependencies, while recent approaches have explored implicit CGs (DICG) \cite{li2021deep}, richer edge modeling \cite{kang2022non,kortvelesy2022qgnn}, and agent grouping strategies~\cite{duan2024group,phan2021vast}. 
These approaches have shown good performance in coordination-intensive environments. However, effective MARL often also requires reasoning about richer and more nuanced interaction dynamics among agents. For instance, different types of dependencies can emerge during learning as agents pursue varying goals, take on complementary roles, or share overlapping capabilities. These may include direct physical interactions (e.g., proximity-based effects), implicit communication links (e.g., signaling intentions or strategies through action choices), or influence-based dependencies (e.g., strategic or cooperative behavior), among others. Furthermore, the effect of an agent’s action can propagate through intermediary agents to influence others. 
Such interaction dynamics are not known a priori and must be discovered and used during learning. Current methods remain limited in their ability to capture any complex and evolving relationships among agents. 

In this paper, we propose a novel algorithm, \algosmall\ (\algoabb), for learning cooperative policies in MARL. It dynamically composes, what we refer to as \textit{meta coordination graphs} (MCGs), which are then used for agent information integration through a graph convolutional network (GCN) \cite{kipf2017semisupervised}. To construct MCGs, it models and learns multiple types of potential relationships among agents that can emerge dynamically over time, while also capturing how information and influence can cascade through intermediate agents. This enhances the integration of information for coordinated decision-making, resulting in a more expressive representation of agent interactions. We term them "meta coordination graphs" because they operate at a higher, \textit{meta}-level of interaction reasoning, i.e., they dynamically integrate richer information about how agents relate and affect one another. 

Instead of explicitly modeling payoffs or utilities as done in standard CG-based methods, \algoabb\ infers how agents influence one another through learned representations. This design offers two key benefits. First, it provides flexible way to encode evolving interaction patterns in cooperative multi-agent tasks through agent information integration. Second, to keep the process fully differentiable, the graphs are jointly optimized with agents’ value functions, enabling end-to-end learning of both expressive interaction representations and coordinated policies. As training progresses, the GCN implicitly reasons about joint actions and values over learned MCGs, allowing them to implicitly guide effective coordination. 

We demonstrate that \algoabb\ achieves superior coordination performance and sample efficiency on multiple tasks that exhibit notable miscoordination challenges. It outperforms both classic and recent CG-based methods, as well as non-graph-based value decomposition MARL baselines. More specifically, our results show that lightweight implicit graphs or simple edge and group modeling can be insufficient, whereas \algoabb\ achieves state-of-the-art performance by reasoning over more complex and diverse agent interactions. 
We also include targeted ablations to quantify the contribution of each component of \algoabb\ and analyze how its gains cannot be matched by merely increasing the parameter count of related CG-based methods. Finally, we also provide a discussion of computational complexity to guide its adoption in scenarios where coordination quality must be balanced against resource constraints.

\section{Related works} 
\label{sec:related}
A popular paradigm in cooperative MARL is \textit{independent learning}, where each agent treats others as part of the environment \cite{tan1993multi,laurent2011world,matignon2012independent}, though it often suffers from non-stationarity and poor coordination. To address this, many approaches adopt \textit{value function factorization}, where the joint value function is represented in terms of individual utilities \cite{rashid2018qmix,sunehag2018value,rashid2020weighted,foerster2018counterfactual,son2019qtran,wang2021qplex}. For example, VDN \cite{sunehag2018value} assumes an additive decomposition of per-agent values, while QMIX \cite{rashid2018qmix} introduces a mixing network that enables monotonic combinations conditioned on the global state. These methods improve scalability and credit assignment but often fail to capture inter-agent dependencies explicitly, leading to miscoordination and issues such as relative overgeneralization \cite{bohmer2020deep,wang2022contextaware,panait2006biasing}.

To address these limitations, \textit{coordination graphs} have been proposed as a structured framework to encode localized interactions among agents \cite{guestrin2002coordinated}. CG-based methods represent agents as nodes and define pairwise payoff functions along edges, supporting structured value decomposition. Deep Coordination Graphs (DCG) \cite{bohmer2020deep} extend this idea to deep MARL by applying a fixed coordination topology and computing value functions over neighboring agents. However, static structures may not generalize well across tasks with dynamic or context-sensitive coordination patterns.

Building on this foundation, recent approaches incorporate more flexible representations of agent interactions. Deep Implicit Coordination Graphs (DICG) \cite{li2021deep} replace explicit edge weights with attention-based soft adjacency matrices and use graph convolution to integrate agent information in a fully differentiable manner. Other works enhance coordination flexibility by learning sparse or context-aware graphs, introducing richer edge representations, or proposing nonlinear graph-based value decompositions \cite{kang2022non,wang2022contextaware}. 

A complementary direction focuses on learning agent \textit{groupings} to simplify the coordination structure. These methods cluster agents into sub-teams and apply factorization within or across groups to reduce complexity and improve scalability \cite{phan2021vast,duan2024group}. By factoring over group-level representations or selectively modeling inter-group dependencies, such approaches offer a middle ground between fully centralized and fully decentralized coordination. 

Our work builds on these developments by incorporating expressive, dynamically composed \emph{meta coordination graphs} to guide information integration among agents, while retaining the structured value function decomposition offered by CG-based methods. By combining these two perspectives, \algoabb\ enables enhanced modeling of interaction dynamics among agents, while still retaining the benefits of structured coordination graphs for scalable credit assignment and modular value decomposition. 

\section{Preliminaries} 
\label{sec:preliminary}
\context{DecPOMDPs.} 
In this paper, we model cooperative multi-agent tasks as a \textit{decentralized partially observable Markov decision process} (Dec-POMDP)~\cite{oliehoek2016concise}, defined by the tuple
\[
\langle I, S, \{A_i\}_{i=1}^{n}, P, R, \{O_i\}_{i=1}^{n}, \Omega, \gamma \rangle,
\]
where \( I = \{1, \ldots, n\} \) is the set of agents and \( S \) is the set of global environment states. Each agent \( i \) has an action space \( A_i \), and the joint action space is \( \boldsymbol{A} = \times_{i \in I} A_i \). At each timestep \( t \), the environment is in state \( s^t \in S \), and the agents take a joint action $a^t = \langle a_1^t, \ldots, a_n^t \rangle \in \boldsymbol{A}$.
The environment transitions to a new state \( s^{t+1} \sim P(\cdot \mid s^t, a^t) \), and all agents receive a shared reward \( r^t = R(s^t, a^t) \). Each agent observes the environment partially via \( o_i^t \in O_i \), drawn from the observation function \( \Omega(o^t \mid s^t) \), where the joint observation is \( o^t = \langle o_1^t, \ldots, o_n^t \rangle \). The joint observation space is \( \boldsymbol{O} = \times_{i \in I} O_i \), and \( \gamma \in [0,1) \) is the discount factor. Each agent maintains a local action-observation history $\tau_i^t = \langle o_i^0, a_i^0, \ldots, o_i^t, a_i^t \rangle$,
and the joint history is \( \tau^t = \langle \tau_1^t, \ldots, \tau_n^t \rangle \). We focus on episodic tasks with horizon \( T \). Let \( \pi_i \) denote the local policy of agent \( i \); the joint policy is
\[
\pi(a^t \mid \tau^t) = \prod_{i \in I} \pi_i(a_i^t \mid \tau_i^t).
\]
The joint action-value function under policy \( \pi \) is
\[
Q_{\pi}(s^t, a^t) = \mathbb{E}_{\pi}\!\left[\sum_{t'=t}^{T} \gamma^{\,t'-t} r^{t'} \,\middle|\, s^t, a^t\right],
\]
and the goal in cooperative MARL is to learn an optimal joint policy \( \pi^* \) that maximizes the expected return, i.e., \( Q^* = \max_{\pi} Q_{\pi} \). 


\paragraph{Value function factorization.} Directly learning the full joint action-value function is intractable, so many MARL methods instead learn to decompose the global joint action-value function \( Q_{\text{tot}}\sim Q^* \) into individual agent value functions \( Q_i \) in different ways while maintaining consistency with the individual greedy maximization (IGM) principle \cite{son2019qtran}. VDNs~\cite{sunehag2018value} adopt a simple additive approach, assuming that the global value can be represented as the sum of individual utilities:
\[
Q^{\text{VDN}}_{\text{tot}}(\tau, a) = \sum_{i=1}^{n} Q_i(\tau_i, a_i),
\]
Whereas, QMIX~\cite{rashid2018qmix} introduces a more expressive architecture in which a mixing network combines the individual utilities via a non-linear monotonic function:
\[
Q^{\text{QMIX}}_{\text{tot}}(\tau, a) = f_{\text{mix}}(Q_1, Q_2, ..., Q_n; s),
\]
where \( f_{\text{mix}} \) is a neural network conditioned on the global state \( s \). 


\paragraph{Coordination Graphs.} CGs~\cite{guestrin2002coordinated} offer a more structured alternative. Rather than fully decomposing \( Q_{\text{tot}} \) into per-agent terms, CGs model localized interactions between agents based on a predefined or learned graph structure. Specifically, for an undirected coordination graph \( G = \langle V, E \rangle \), where each node \( v_i \in V \) corresponds to an agent and each edge \( \{i, j\} \in E \) encodes a coordination dependency between agents \( i \) and \( j \), the joint action-value can be factorized as:
\[
Q^{\text{CG}}_{\text{tot}}(\tau, a) = \frac{1}{|V|} \sum_{v_i \in V} Q_i(\tau_i, a_i) + \frac{1}{|E|} \sum_{\{i,j\} \in E} Q_{ij}(\tau_i, \tau_j, a_i, a_j),
\]
where \( Q_i \) represents the individual utility function of agent \( i \), and \( Q_{ij} \) captures the pairwise payoff between agents \( i \) and \( j \). 

\paragraph{Implicit Coordination Graphs.} Some methods (e.g., DICG) replace explicit coordination graph factorization with learned, implicit interaction structures parameterized by attention and graph neural networks (GNNs). Instead of learning binary edge weights in a coordination graph, one can compute a soft adjacency matrix \(M \in \mathbb{R}_{>0}^{n \times n}\) using self-attention over encoded agent observations \(\{e_i\}_{i=1}^n\):
\[
\mu_{ij} = 
\frac{\exp(\text{Attention}(e_i, e_j; W_a))}{\sum_{k=1}^{n}\exp(\text{Attention}(e_i, e_k; W_a))},
\quad
M_{ij} = \mu_{ij},
\]
where \(W_a\) is a trainable weight matrix. This soft adjacency captures the strength of influence between agents and forms an implicit coordination graph. A feature matrix \(E^{(0)} = [e_1^\top; \dots; e_n^\top]\) can then be updated by graph convolutional layers to integrate information across agents:
\[
H^{(\ell+1)} = \sigma\big(MH^{(\ell)}W_c^{(\ell)}\big), 
\quad H^{(0)} = E^{(0)},
\]
where \(H^{(\ell)} \in \mathbb{R}^{n \times d_\ell}\) are the agent embeddings at layer \(\ell\), \(W_c^{(\ell)}\) are trainable weights and \(\sigma\) is a non-linear activation. Stacking multiple layers yields integrated agent embeddings \(E^{(m)}\) that aggregate contextual information from neighbors. These GNN-based embeddings can be used directly for policy or value estimation, effectively bypassing the traditional coordination graph decomposition.

\section{\algotitle} 
\label{sec:methodology}

\begin{figure*}[!ht]
    \centering
    \includegraphics[width=\linewidth]{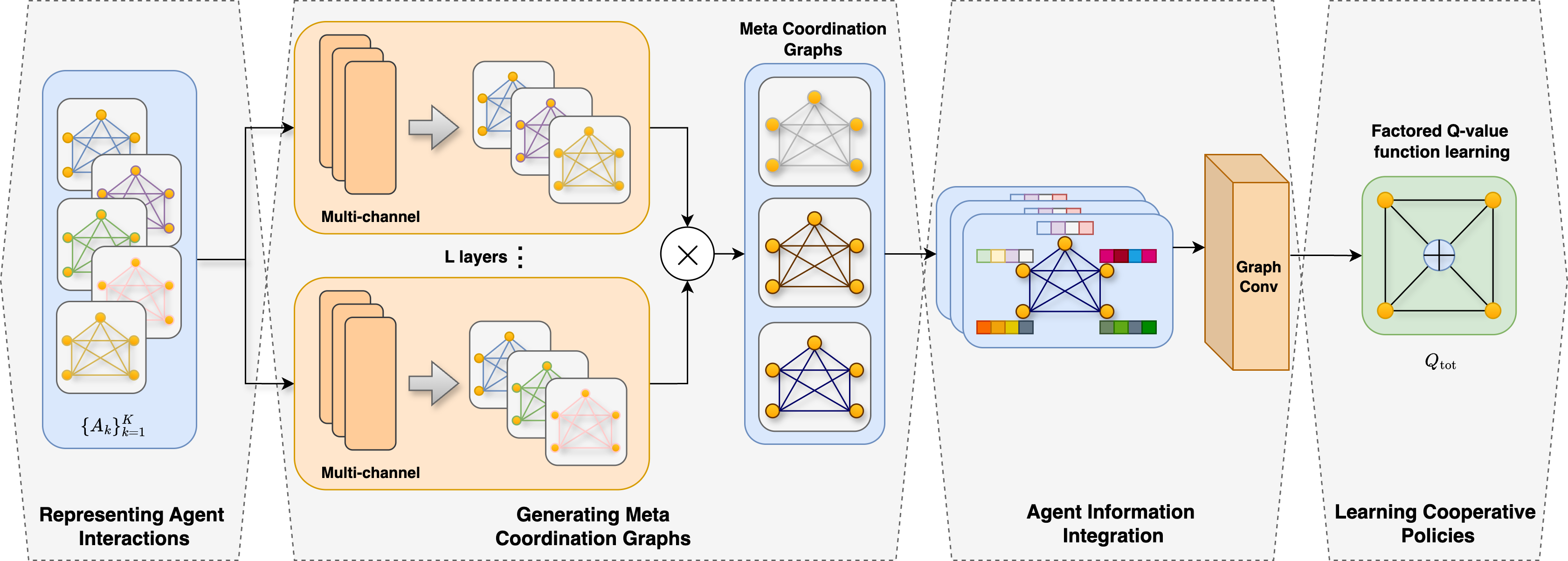} 
    \caption{Overview of \algotitle\ (\algoabb). From left to right: (1) agent interactions are represented through a set of base relation graphs, (2) multiple attention-based composition layers combine these into task-adaptive \textit{meta coordination graphs} (MCGs), (3) MCGs guide graph convolutions for agent information integration, and (4) the resulting embeddings are used for factored Q-value learning to produce cooperative policies. Together, these steps enable \algoabb\ MARL agents to achieve robust coordination and strong sample efficiency across challenging tasks.} 
    \label{fig:methodology}
\end{figure*}

In this section, we present \algosmall\ (\algoabb), designed to improve coordination by learning how agents should exchange and combine information. The goal is to let each agent make better decisions by selectively using signals from others and adapting to how their interactions evolve over time. We define an \emph{interaction} as a latent dependency through which one agent's observations or actions can influence another agent’s choices. To model emerging dependencies, \algoabb\ starts with a set of base graphs with pairwise edges. 
These are then softly reweighted and combined using attention to create \emph{meta coordination graphs} (MCGs), which adapt as training progresses and as the task reveals how agents actually interact with one another. Next, graph convolutions over these MCGs propagate and integrate information among agents through message-passing, building expressive representations for each agent’s decision-making. Finally, both the learned MCGs and the value function factorization are optimized jointly in a fully differentiable, end-to-end pipeline. This design lets \algoabb\ discover and use rich, underlying coordination dynamics while keeping the benefits of graph-based MARL. Figure~\ref{fig:methodology} summarizes the overall architecture and learning workflow. 

\paragraph{Representing agent interactions.} In particular, we represent the multi-agent system as an undirected graph \(\mathcal{G}=(\mathcal{V},\mathcal{E})\),  
where the node set \(\mathcal{V}=\{1,\dots,n\}\) corresponds to the \(n\) agents and  
the edge set \(\mathcal{E}\subseteq \mathcal{V}\times\mathcal{V}\) represents potential influence between agents.  
To capture multiple potential interactions, we maintain a set of \(K\) \emph{base relation graphs},  
each stored as an adjacency matrix \(A_k\in\mathbb{R}^{n\times n}\).  
These matrices form a third-order tensor
\[
\bm{A}=\{A_k\}_{k=1}^{K}.
\]
Each \(A_k\) corresponds to a distinct latent relation type \(e_k\in\mathcal{E}\),  
and the entry \(A_k[i,j]\) indicates the potential influence of agent \(i\) on agent \(j\) under relation \(e_k\). All base graphs are initialized as complete graphs
serving as a neutral, maximally expressive starting point from which the model can later learn to select and combine relation types through attention. The number of base graphs \(K\) controls the diversity of latent interaction types available for modeling. A large \(K\) expands the search space of possible relations but also increases computational cost. We analyze this trade-off in more detail later (Section~\ref{sec:results}). Each agent \(i\) contributes a local observation \(o_i \in \mathbb{R}^{d}\), and these form the observation or fetaure matrix \(X = [o_1; \ldots; o_n] \in \mathbb{R}^{n \times d}\). Together, \(\bm{A}\) and \(X\) define the structured input space from which the model constructs task-adaptive MCGs in the next stage. 

\paragraph{Generating meta coordination graphs.} Next, \algoabb\ constructs MCGs through a sequence of $L$ composition layers. Each composition layer learns to mix the base graphs in a channel-wise manner. $C$ parallel channels are constructed, each using a different learnable mixture of the same $K$ base graphs.  
Formally, at each composition layer $\ell \in \{1, \dots, L\}$, the model computes a soft mixture of the $K$ base graphs for each channel $c \in \{1, \dots, C\}$ as:
\[
A^{(l,c)}=\sum_{k=1}^{K}\alpha_k^{(l,c)}A_k 
\]
where $\alpha_k^{(\ell, c)}$ denotes the soft attention weight assigned to the $k^{\text{th}}$ base graph at layer $\ell$ and channel $c$. These weights are learned independently for each channel and layer as trainable scores over the $K$ base graphs. 
The attention mechanism is parameterized by a layer-specific matrix \( W^{(\ell)} \in \mathbb{R}^{C \times K} \), where each row \( W^{(\ell)}_c \) defines the attention logits over the \( K \) base graphs for channel \( c \). 
After \(L\) such layers, the adjacency for channel \(c\) is obtained by sequential matrix multiplication:
\[
A_M^{(c)}=\prod_{\ell=1}^{L}A^{(\ell,c)} 
\]
This yields \(C\) final adjacency matrices (i.e., MCGs) \(\{A_M^{(c)}\}_{c=1}^{C}\), each representing a learned pattern of how agents may influence one another. This composition mechanism serves two key purposes. First, the \(C\) parallel channels allow the model to maintain multiple plausible hypotheses about the underlying coordination dynamics, capturing diverse and potentially complementary modes of agent interaction. Second, the sequential matrix multiplications across \(L\) layers implicitly enable \algoabb\ to capture multi-hop relational dependencies by softly weighting and composing different base relations at each layer. This soft mixing favors certain interaction patterns while down-weighting others, resulting in multi-step influence chains without the need to explicitly model long-range edges. The final MCGs are then used to guide agent information integration in the subsequent graph convolution layers. 

\paragraph{Agent information integration.} The MCGs \(\{A_M^{(c)}\}_{c=1}^{C}\) are then used to propagate and aggregate information among agents. Each weighted graph \(A_M^{(c)}\) acts as an independent message-passing channel, allowing the model to process each MCG in parallel. To ensure agents retain access to their own embeddings during message passing, self-loops are explicitly added. We use the standard form: $\tilde{A}_M^{(c)} = A_M^{(c)} + I$. Thus, for each channel \(c\), we have
\[
H^{(c)}=\sigma\!\left(\tilde{D}_c^{-1}\tilde{A}_M^{(c)}XW\right),
\]
where \(\tilde{D}_c\) is the degree matrix of \(\tilde{A}_M^{(c)},\) \(X\in\mathbb{R}^{n\times d}\) is the matrix of agent features, and \(W\in\mathbb{R}^{d \times d_{\mathrm{emb}}}\) is a shared trainable weight matrix, with \(d_{\mathrm{emb}}\) denoting the dimensionality of the latent embedding space used to represent each agent after message passing. Each \(H^{(c)}\in\mathbb{R}^{n\times d_{\mathrm{emb}}}\) contains the updated representations produced by message passing over graph \(A_M^{(c)}\). The outputs from all channels are concatenated along the feature dimension,
\[
Z=\big\|_{c=1}^{C}H^{(c)}\in\mathbb{R}^{n\times (C\,d_{\mathrm{emb}})},
\]
so that each agent’s representation integrates information aggregated under multiple, task-adaptive relation types. A final linear projection maps \(Z\) back to the embedding size \(d_{\mathrm{emb}}\), yielding per-agent features that combine information from all \(C\) meta graphs. 

\paragraph{Learning cooperative policies.} The final agent embeddings \(Z \in \mathbb{R}^{n \times (C\,d_{\mathrm{emb}})}\), obtained after MCG-guided information integration, replace the raw local inputs in each agent’s trajectory. As a result, policies and value estimators reason over information that has already been aggregated from other agents. To estimate joint action values, we retain the CG-based value factorization, which is effective for coordination-intensive tasks. Following the deep coordination graph (DCG) formulation~\cite{bohmer2020deep}, the global action-value function is decomposed as
\[
\begin{split}
Q(\tau^t, a)
=\;& \frac{1}{|\mathcal{V}|} \sum_{i} Q_i(a_i \mid \tau^t_i) \\
&+ \frac{1}{2|\mathcal{E}|} \sum_{(i,j)\in\mathcal{E}}
\bigl[Q_{ij}(a_i,a_j\mid\tau^t_i,\tau^t_j)
+Q_{ji}(a_j,a_i\mid\tau^t_j,\tau^t_i)\bigr],
\end{split}
\]
where each local trajectory \(\tau^t_i\) now contains the MCG-enhanced embedding \(z_i\) rather than the raw observation \(o_i\). Here, \(\mathcal{E}\) is replaced with the complete CG as in DCG~\cite{bohmer2020deep}, while the enhanced representations implicitly guide cooperative policy learning. Directly replacing \(\mathcal{E}\) with a dynamically evolving MCG could break the theoretical guarantees of CG-based value factorization, but \algoabb\ allows for using MCG embeddings to capture dynamic interaction patterns in a fully differentiable end-to-end pipeline.

\paragraph{Key distinctive aspects of this approach.} \algoabb\ develops upon several foundational ideas in representation learning and graph modeling, and applies them in a novel way to cooperative MARL. Its use of multiple relation channels is similar to channel attention and feature pooling in computer vision~\cite{chen20182}, where different channels capture complementary patterns and are adaptively weighted. Whereas, \algoabb\ maintains a bank of \(K\) latent relation types \(\{A_k\}_{k=1}^{K}\) and learns attention weights over them to form relation channels. Their sequential composition across \(L\) layers connects to work on neural relation composition and meta-path reasoning~\cite{wang2019heterogeneous,zhang2018deep,yun2019graph}, where adjacency multiplication captures higher-order dependencies. This is integrated with standard graph convolutional message passing~\cite{kipf2017semisupervised}, widely used in MARL~\cite{munikoti2023challenges,chen2021graph}, but enhanced here to reason over evolving relational contexts. These ideas set \algoabb\ apart from prior approaches. Edge-weighting approaches such as DICG~\cite{li2021deep} apply attention directly to individual edges of a coordination graph, producing a soft adjacency matrix. And, grouping-based methods such as VAST~\cite{phan2021vast} and GACG~\cite{duan2024group} explicitly cluster agents into predefined or learned groups and model interactions between or within those groups.

\section{Experiments} 
\label{sec:experiments}

In this section, we present the setup for empirical evaluation of our proposed approach, \algoabb, in various MARL tasks. 

\subsection{Evaluation domains} 


We evaluate \algoabb\ on four environments from the Multi-Agent Coordination (MACO) benchmark~\cite{wang2022contextaware}. It was introduced to stress-test algorithms on tasks where coordination is the primary difficulty rather than low-level control. It builds on several classic cooperative problems but deliberately increases their complexity by introducing temporal extension, stochastic dynamics, and larger agent teams. 
MACO focuses on scenarios where agents must reason about collective behavior, dynamically changing dependencies, and penalties for miscoordination. This makes it well-suited to evaluating algorithms designed to model rich interactions and adapt coordination strategies. We summarize the four tasks (Gather, Disperse, Pursuit, and Hallway) used in our experiments below and provide further details on reward definitions and implementations in the appendix. 

\begin{figure}[!htp]
    \centering
    \captionsetup[subfigure]{justification=centering}
    \subfloat[Gather]{\includegraphics[width=0.3\linewidth]{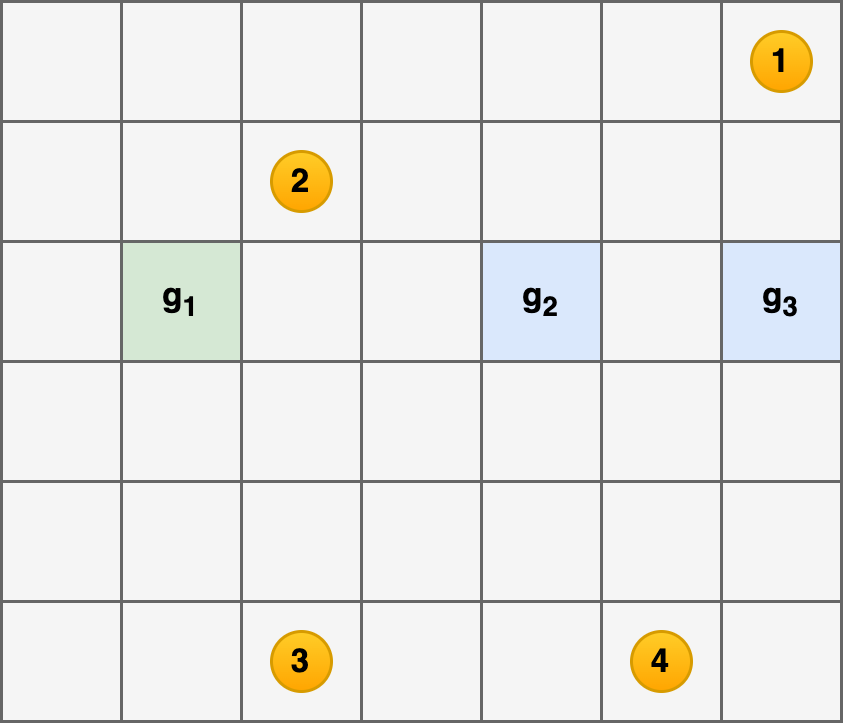}} \qquad
    \subfloat[Disperse]{\includegraphics[width=0.6\linewidth]{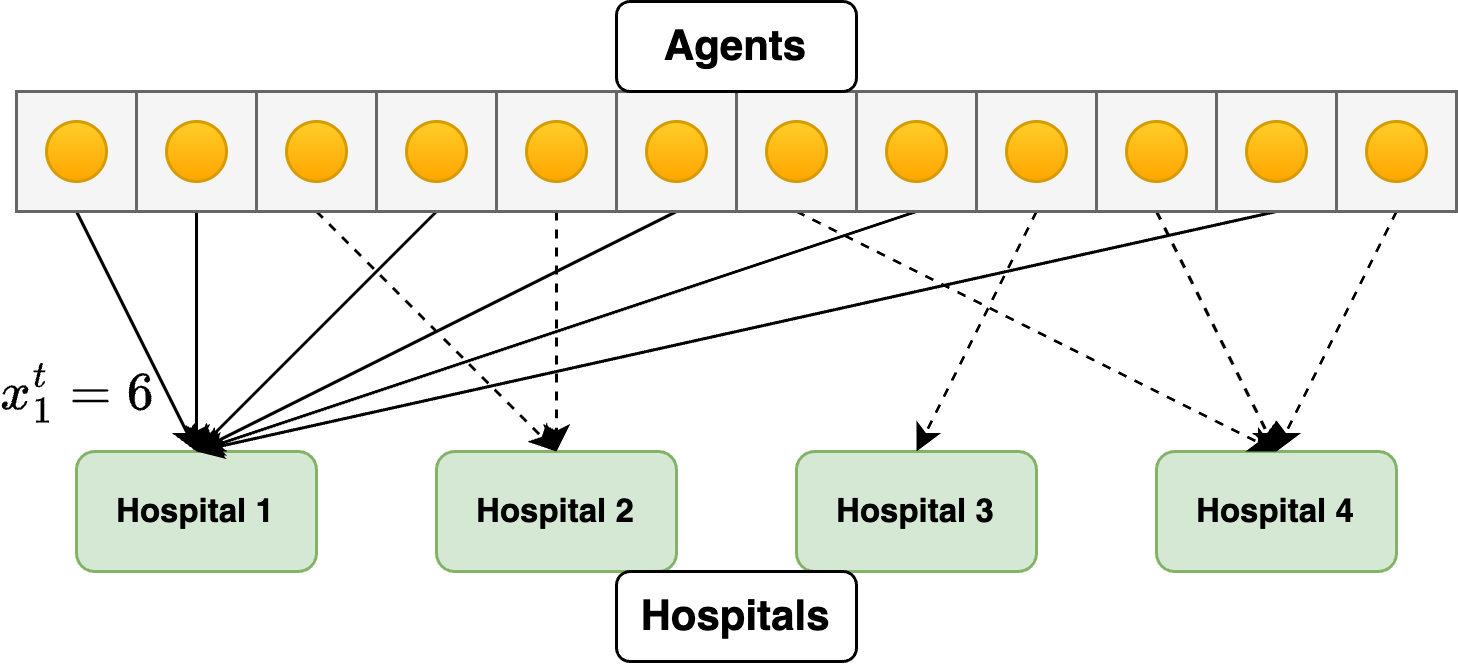}}  \\
    \subfloat[Pursuit]{\includegraphics[width=0.3\linewidth]{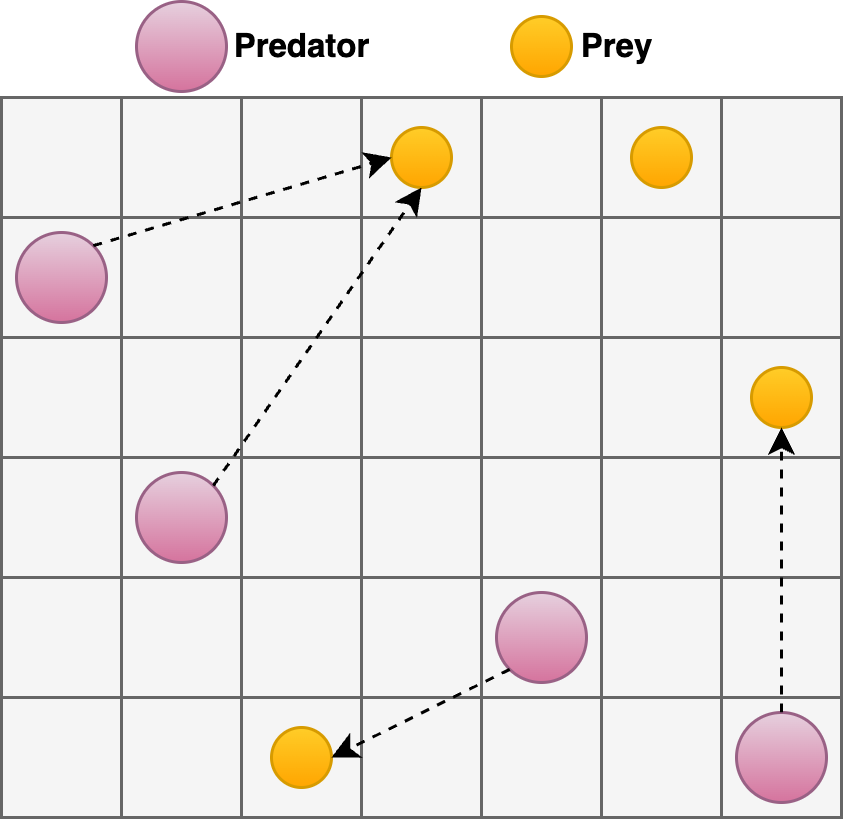}} \qquad
    \subfloat[Hallway]{\includegraphics[width=0.6\linewidth]{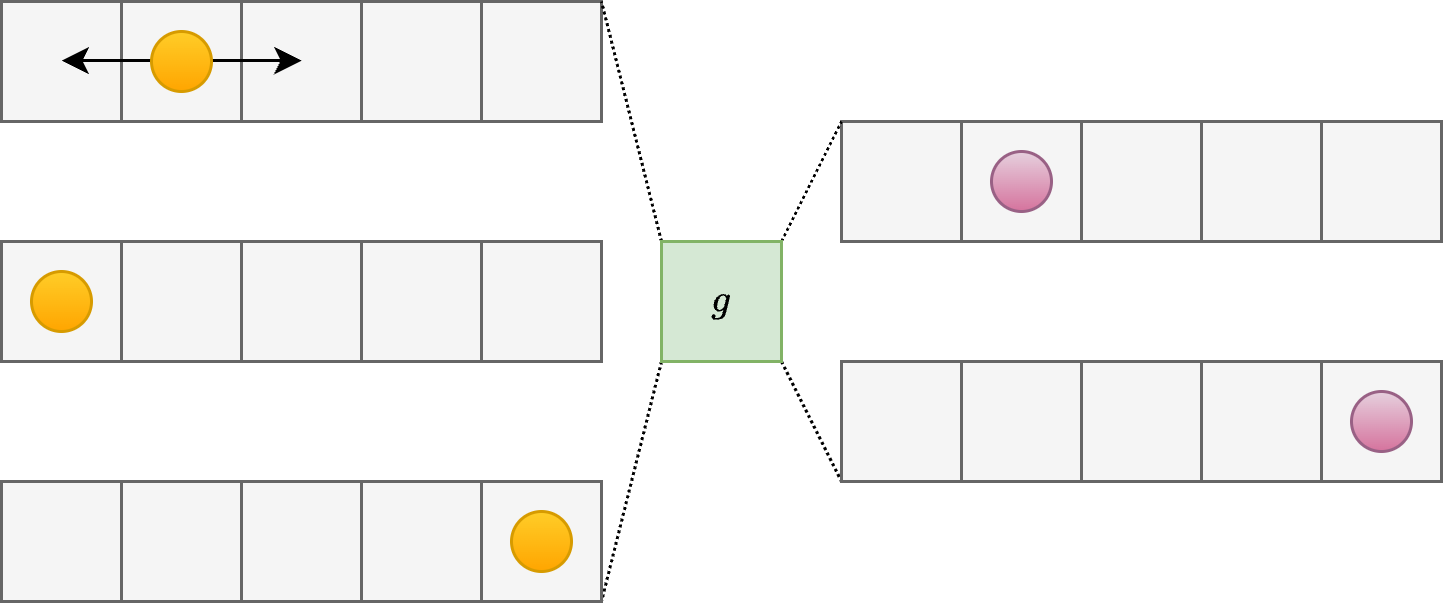}} 
    \caption{Evaluation environments: (a) Gather with optimal goal $g_1$, (b) Disperse requiring 6 agents at Hospital 1, (c) Pursuit where 2 predators (purple) must capture a prey (yellow), and (d) Hallway with 2 agent groups.}
    \label{fig:envs}
\end{figure} 

\textbf{Gather} (Figure~\ref{fig:envs}a) is a temporally extended and partially observable variant of the classic Climb Game~\cite{wei2016lenient}, involving five agents. Each episode randomly designates one of three goal locations (\(g_1, g_2, g_3\)) as the optimal target, but only agents spawned nearby know which goal is correct. Agents must navigate and simultaneously arrive at the same goal to earn a team reward. The reward is high if all reach the optimal goal, moderate if all choose a suboptimal goal, and severe if they split. 
\textbf{Disperse} (Figure~\ref{fig:envs}b) features twelve agents repeatedly choosing among four hospitals, where only one hospital requires a nonzero and changing number of agents each timestep. Agents are penalized for under- or over-staffing, forcing them to adaptively redistribute based on limited and dynamically changing local information. \textbf{Pursuit} (Figure~\ref{fig:envs}c) is a challenging variant of the classic predator-prey game~\cite{stone2000multiagent}. Ten predator agents move on a \(10 \times 10\) grid to capture randomly moving prey. A capture succeeds only if at least two predators coordinate their actions simultaneously on the same prey; failed attempts are penalized, making precise temporal coordination essential. \textbf{Hallway} (Figure~\ref{fig:envs}d) extends the multi-chain Dec-POMDP of~\cite{wang2019learning} by increasing the number of agents and introducing conflicting group dynamics. Agents must synchronize movements through constrained corridors and reach a shared goal state simultaneously, while avoiding conflicts when multiple groups attempt to pass at once under partial observability.

\subsection{Learning Algorithms and Evaluations} 

To assess \algoabb, we compare it against a broad and representative set of MARL baselines spanning different coordination paradigms. \textbf{VDN} \cite{sunehag2018value} and \textbf{QMIX} \cite{rashid2018qmix} represent popular value factorization methods that decompose the global value function without explicitly modeling structured interactions. We include several graph-based approaches: \textbf{DCG} \cite{bohmer2020deep} uses a static, predefined pairwise graph; \textbf{DICG} \cite{li2021deep} learns a single implicit graph dynamically via attention; \textbf{CASEC} \cite{wang2022contextaware} builds sparse, context-aware graphs using payoff variance; and \textbf{NLCG} \cite{kang2022non} applies non-linear mixing of graph-based utilities. We also consider subgrouping-based methods: \textbf{GACG} \cite{duan2024group} forms adaptive agent groups and \textbf{VAST} \cite{phan2021vast} uses hierarchical subgroup value decomposition. These baselines together cover the major families of value decomposition methods in cooperative MARL: full value factorization, graph-based coordination, and group-based coordination. 
Further implementation and training details, including hyperparameters and environment-specific reward formulations, are also provided in the appendix. To ensure statistical significance, we run each algorithm with four independent seeds in every environment and report the mean and standard deviation of the relevant performance metric (e.g., win rate, return, prey captured).

\section{Results and Discussion} 
\label{sec:results}
We now evaluate \algoabb\ on the aforementioned cooperative MARL tasks and present a comprehensive analysis of its performance and design. Our aim is to answer three key questions: (1)~does \algoabb\ outperform existing value-factorization and coordination-graph methods on challenging multi-agent tasks; (2)~which components of its design are essential for strong and stable learning; and (3)~how its added expressivity and flexibility balance against computational cost in practice. Our study first establishes \algoabb's overall effectiveness compared to a diverse set of state-of-the-art baselines, and then uses controlled ablations and targeted analyses to isolate the benefit of its MCG-driven agent information integration. 
We then also examine the effect of dense versus sparse initial base graphs and conclude with a discussion on the computational complexity and runtime trade-offs of this approach. 

\begin{figure*}[!htp]
    \centering
    \includegraphics[width=\linewidth]{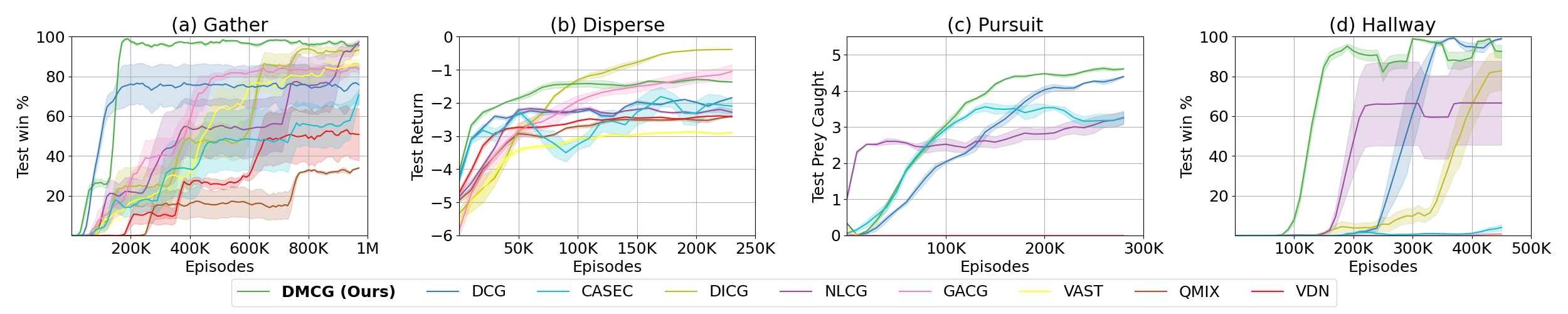} 
    \caption{\textbf{Overall performance on MACO benchmark.} Comparison of \algoabb\ with representative MARL baselines across four tasks: \textit{Gather}, \textit{Disperse}, \textit{Pursuit}, and \textit{Hallway}. \algoabb\ consistently achieves state-of-the-art or near-optimal performance, converging faster and attaining higher mean episode returns than static graph (DCG), attention-based graph (DICG), edge-selection (CASEC), non-linear mixers (NLCG), subgrouping (GACG, VAST), and value-decomposition methods (QMIX, VDN).}
    \label{fig:res:maco_sota}
\end{figure*}

\paragraph{Main results.} Figure~\ref{fig:res:maco_sota} compares \algoabb\ against other MARL baselines on the MACO benchmark. For our main experiments, we set the number of base relations $K$, channels $C$, and composition layers $L$ in \algoabb\ equal to the number of agents in each environment. A bank of $K$ fully connected base graphs offers a rich initial set of candidate interaction patterns; $C$ parallel channels let the model form diverse soft mixtures of these bases; and setting $L$ to the agent count provides enough depth to compose them through successive matrix multiplications, enabling long coordination chains while remaining a practical upper bound. We also validate and analyze these choices through targeted ablations later. 

\algoabb\ consistently achieves the best or near-best performance across all four environments. In \textbf{Gather} (Fig.~\ref{fig:res:maco_sota}a), \algoabb\ rapidly achieves near-optimal win rates ($\sim$98--100\%) within $\approx$200K episodes, significantly surpassing all baselines. DCG plateaus at suboptimal levels ($\sim$80\%), while DICG improves steadily yet remains below \algoabb\ throughout training. Subgrouping methods (GACG, VAST) show competitive learning later (after $\approx$650K episodes) but settle at lower returns. CASEC and NLCG also lag throughout the training, and fully factorized value-decomposition baselines (QMIX, VDN) fail to learn meaningful coordination within the 1M-episode training regime for Gather. In \textbf{Disperse} (Fig.~\ref{fig:res:maco_sota}b), while DICG eventually attains the highest final score and GACG narrows the gap, \algoabb\ learns markedly faster and reaches competitive performance with far fewer episodes. This can be a crucial advantage in domains where data collection is costly or training time is limited. \algoabb\ outperforms all other baselines. On the more challenging \textbf{Pursuit} (Fig.\ref{fig:res:maco_sota}c) and \textbf{Hallway} (Fig.\ref{fig:res:maco_sota}d), where coordination failure leads to severe penalties (e.g., predators failing to jointly capture prey, or agents colliding in partially observable hallways), \algoabb\ again achieves clear and consistent advantages. It surpasses DCG and DICG in both convergence speed and final task score, while fully factorized methods fail to learn meaningful policies. CASEC (in Pursuit) and VAST (in Hallway) perform better than QMIX and VDN but remain below \algoabb. NLCG exhibits reasonable but suboptimal performance in both tasks, whereas GACG struggles to learn effective strategies. 

Overall, the results show that \algoabb\ reliably outperforms fully factorized value-decomposition baselines (QMIX, VDN) and achieves stronger or comparable performance than recent graph-based methods, including static CGs (DCG), attention-based implicit CGs (DICG), richer edge modeling (CASEC), non-linear CGs (NLCG), and subgrouping approaches (GACG, VAST), across the evaluated multi-agent tasks. For all baselines, we adopt widely used and actively maintained public implementations with their standard recommended hyperparameters. DCG is run with a complete graph and no low-rank optimization. DICG follows its recommended setup with two GCN layers. NLCG uses its non-linear edge mixing with a 3-dimensional hidden mixing network. CASEC is configured with payoff-variance-driven edge construction (threshold = 0.3). GACG uses two learned groups and two GCN layers. VAST adopts its hierarchical subgrouping with two groups. Non-graph baselines (QMIX, VDN) follow typical recurrent agent settings. 

After confirming \algoabb's overall performance gains, we next analyze its design choices through a series of controlled studies on the \textit{Gather} environment, which serves as a consistent, coordination-intensive benchmark for the following experiments. 

\begin{figure}[!htp]
    \centering
    \includegraphics[width=0.65\linewidth]{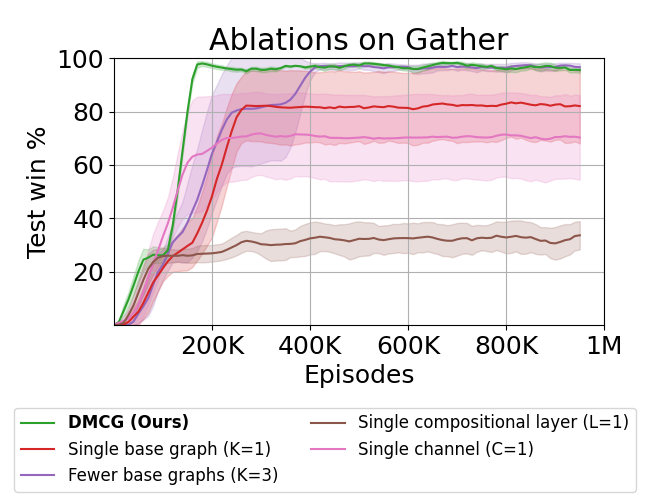} 
    \caption{\textbf{Targeted ablations of \algoabb.} Studying the effect of varying composition depth ($L$), number of channels ($C$), and base relation bank size ($K$) in \algoabb.} 
    \label{fig:res:gather_ablations}
\end{figure}

\paragraph{Ablations.} To understand which design choices make \algoabb\ effective, we selectively simplify its three core components on the \textit{Gather} task (Figure~\ref{fig:res:gather_ablations}): the depth of composition layers ($L$), the number of parallel channels ($C$), and the size of the base relation bank ($K$). We focus on the most informative reductions, setting each to its simplest non-trivial form, since these provide a clear contrast without introducing unnecessary variants. 

\textbf{(1) Composition layers ($L$).} We compare the full multi-layer design with a single compositional layer (\textit{$L=1$}). This isolates whether stacking layers is beneficial beyond a simple one-shot combination of base relations. Results show a sharp performance drop with $L=1$, confirming that progressive reweighting and multiplication across layers is critical; a single layer forces the model to infer all dependencies in one step and limits expressiveness. 

\textbf{(2) Parallel channels ($C$).} We test a single channel (\textit{$C=1$}) against the multi-channel default. This ablation investigates if \algoabb\ truly benefits from exploring multiple MCGs in parallel for coordination or not. Performance remains reasonable with $C=1$ but converges to a lower final win rate, showing that multiple channels help the network consider and combine alternative interaction patterns. 

\textbf{(3) Number of base relations ($K$).} We study small base banks, $K=1$ (\textit{Single base graph}) and $K=\lceil n/2\rceil$ (\textit{Fewer base graphs}), against the full $K=n$ setup. These settings expose the trade-off between expressive capacity and cost: $K=1$ forces all coordination through a single relation, $K\approx n/2$ tests a moderate-sized (halved) relation set, and $K=n$ matches the maximum number of agents. Performance improves with larger $K$, showing that a richer relation bank expands the search space for task-relevant dependencies while a very small $K$ limits diversity and sample efficiency.

These studies show that (i) multiple layers help progressively refine interaction structure, (ii) parallel channels allow the model to consider alternative coordination patterns, and (iii) a richer base relation bank expands the search space for discovering task-relevant dependencies. Together, these elements explain \algoabb’s stable and effective performance.

\begin{figure}[!htp]
    \centering
    \includegraphics[width=0.65\linewidth]{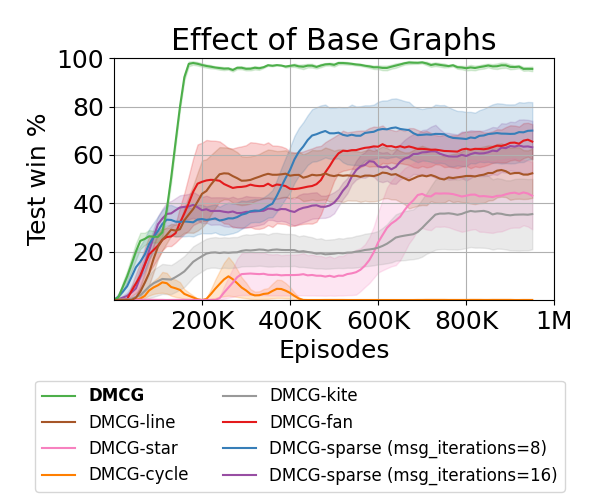} 
    \caption{\textbf{Effect of base graph initialization.} Studying how starting with fully connected vs.\ sparse base graphs affects \algoabb\ on \textit{Gather}. Sparse topologies yield comparatively lower final performance, and adding more message-passing steps (8→16) also does not close the gap.} 
    \label{fig:res:gather_sparse}
\end{figure}

\paragraph{Effect of initialization of base graphs.} We further study how the choice of initial base graphs impacts \algoabb’s ability to discover effective coordination patterns (Figure~\ref{fig:res:gather_sparse}). Our proposed design for \algoabb\ initializes each base relation as a fully connected graph to maximize initial expressivity. To evaluate robustness under sparser priors, we test two structured alternatives: \textbf{\algoabb-same}, where all bases share the same sparse template (we separately tried \textit{line}, \textit{star}, \textit{cycle}, and \textit{kite} topologies, using the same one for every base), and \textbf{\algoabb-diverse}, where the base bank combines one of each of the topologies \{\textit{line}, \textit{star}, \textit{cycle}, \textit{kite}, \textit{full}\}. These variants probe whether \algoabb\ can still form strong MCGs when initial connectivity is reduced, either uniformly or with some diversity. More details on the topologies are in the appendix. As shown in Figure~\ref{fig:res:gather_sparse}, \algoabb-diverse\ performs reasonably and clearly improves over \algoabb-same, indicating that diversity among sparse bases provides richer relational cues and improves adaptation. However, both sparse setups remain below \algoabb. This suggests a practical trade-off: while sparsity can save computation, it constrains the final performance of \algoabb. For applications where peak accuracy is less critical but computational cost is a concern, diverse sparse setups may offer an acceptable compromise. Nonetheless, fully connected initialization remains the most reliable and expressive default when no strong prior structure is known. 

For further analysis, given that message passing is the primary mechanism for propagating influence in graph-based methods, we tested whether increasing the number of message passing iterations (from 8 to 16) could compensate for the reduced performance. A message-passing iteration refers to one round in which each agent aggregates the current messages from its neighbors and updates its own local $Q$-value estimate; repeating this process multiple times allows information to propagate over progressively longer paths in the graph \cite{guestrin2002coordinated,bohmer2020deep}. Prior work on DCGs \cite{bohmer2020deep} suggests that additional iterations can in principle propagate information over longer paths, so we expected more iterations to help sparse variants approximate the expressive power of dense ones. However, we observed no improvement and even a slight drop in performance (Figure~\ref{fig:res:gather_sparse}). We suspect that when the initial graph is already limited, extra message steps mainly amplify noise or redundant signals rather than providing meaningful new information, as the model lacks a rich set of candidate relations to refine. These findings suggest a promising avenue for future work: designing training-time sparsification or pruning strategies that reduce overhead while retaining key relational diversity, instead of enforcing sparsity from the start. 

\begin{figure}[!htp]
    \centering
    \includegraphics[width=0.65\linewidth]{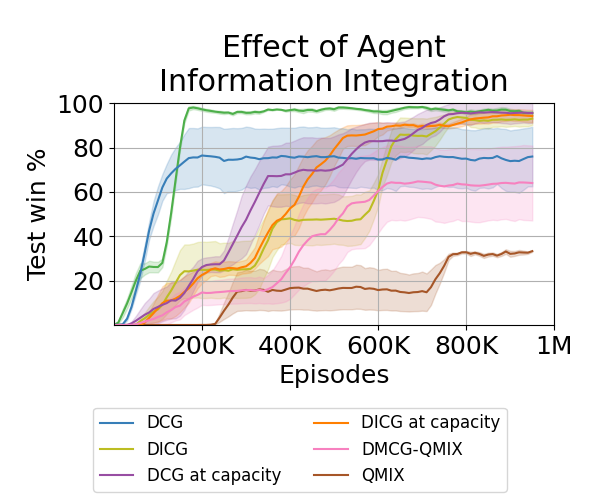} 
    \caption{\textbf{Effectiveness of agent information integration.} Scaling DCG (wider payoffs) and DICG (deeper GCNs) still underperforms \algoabb, showing that gains come from MCG-driven integration rather than larger models. Even QMIX, when extended with DMCG integration, improves notably.} 
    \label{fig:res:gather_overhead}
\end{figure}

\paragraph{Effectiveness of \algoabb's agent information integration.} Interpreting \algoabb’s learned representations directly is challenging. Its base relations, attention weights, and multi-layer compositions operate in a latent space, making qualitative inspection of the discovered coordination patterns difficult. To qualitatively assess and isolate the benefit of this MCG-driven agent information integration, we compare \algoabb\ against capacity-scaled variants of strong baselines (Figure~\ref{fig:res:gather_overhead}). 
We increase DCG's capacity by doubling the hidden size of its payoff networks (64\,$\rightarrow$\,128) and enlarge DICG by extending its number of GCN layers (2\,$\rightarrow$\,5 layers), scaling both baselines to have a parameter count approximately comparable to \algoabb. While these upgrades yield moderate gains for DCG and DICG, their final performance remains below \algoabb, indicating that simply adding more learnable parameters to learn from static graphs or deepening attention-based edge weighting cannot match the expressive agent information integration enabled by MCG composition. We further validate this by implementing \algoabb’s information-integration module over QMIX, which does not encode any structured inter-agent reasoning. We observe clear improvements over vanilla QMIX. Together, these results confirm that \algoabb’s advantage comes from its compositional agent information integration rather than simply larger networks or deeper conventional message passing.

\paragraph{Complexity and runtime analysis.} The computational cost of \algoabb\ scales with the number of agents $n$, the number of base relations $K$, the number of channels $C$, the number of composition layers $L$, and the embedding dimension $d_{\mathrm{emb}}$. For dense graphs, each composition layer forms $C$ soft attention mixtures over the $K$ base adjacency matrices, requiring $O(K\,C\,n^{2})$ operations, and then sequentially multiplies these $n \times n$ matrices across $L$ layers, adding $O(L\,C\,n^{3})$. The resulting $C$ MCGs are each processed by a graph convolution, which costs $O(C\,n\,d_{\mathrm{emb}}^{2}+C\,n^{2}d_{\mathrm{emb}})$. Overall, a forward pass scales as
\[
O(L\,K\,C\,n^{2} + L\,C\,n^{3} + C\,n\,d_{\mathrm{emb}}^{2} + C\,n^{2}d_{\mathrm{emb}}),
\]
where the $L\,C\,n^{3}$ term from sequential matrix multiplications and the $C\,n^{2}d_{\mathrm{emb}}$ term from GCN aggregation typically dominate, thus giving an approximate leading cost of $O(L\,C\,n^{3}+C\,n^{2}d_{\mathrm{emb}})$. While this theoretical complexity appears high, the dominant operations can be reduced to large batched matrix multiplications that are highly parallelizable and well-optimized on modern GPUs. 

\begin{table}[h]
\centering
\caption{Comparison of computational complexity and performance. Each cell shows final performance / episodes (K, to reach peak performance) / wall-clock time (hrs).}
\label{tab:complexity_comparison}
\resizebox{\linewidth}{!}{%
\begin{tabular}{lcccc}
\toprule
\textbf{Algorithm} & \textbf{Gather} & \textbf{Disperse} & \textbf{Pursuit} & \textbf{Hallway} \\
\midrule
\textbf{DMCG}   & 98\% / 180K / 9h   & -1.4 / 100K / 15h  & 4.6 / 200K / 24h   & 99\% / 300K / 30h \\
\textbf{DICG}   & 97\% / 800K / 2.5h & -0.4 / 200K / 1h   & 0.0 / 300K / 8h    & 82\% / 450K / 3h  \\
\textbf{DCG}    & 78\% / 200K / 1.5h & -2.0 / 150K / 5h   & 4.4 / 300K / 12h   & 98\% / 325K / 14h \\
\textbf{CASEC}  & 67\% / 1000K / 5h  & -1.8 / 175K / 1.5h & 3.5 / 225K / 4h    & 5\% / 450K / 5h   \\
\textbf{NLCG}   & 99\% / 1000K / 6h       & -2.2 / 50K / 4h       & 3 / 300K / 7h       & 65\% / 450K / 5h      \\
\textbf{GACG}   & 83\% / 650K / 3h       & -1.0 / 250K / 2h       & 0 / 300K / 3h       & 2\% / 450K / 3h      \\
\textbf{VAST}   & 85\% / 850K / 2h       & -3.0 / 150K / 1h       & 0 / 300K / 1h       & 0\% / 450K / 2h      \\
\bottomrule
\end{tabular}
}
\end{table} 

Empirically, Table~\ref{tab:complexity_comparison} reports wall-clock training time, sample efficiency (episodes to reach peak performance), and final task scores for major CG-based methods. While \algoabb\ incurs higher per-step computation due to its compositional attention, it typically requires far fewer episodes to converge and achieves the best or near-best task performance. For example, in \textit{Gather}, \algoabb\ reaches $98\%$ win rate in $\sim$180K episodes (9h) versus DCG’s $78\%$ at 200K (1.5h) and CASEC’s $67\%$ after 1M (5h). In \textit{Pursuit}, \algoabb\ also converges faster in episode count and attains the highest prey-capture score, despite longer wall-clock time per step. This reflects a trade-off: \algoabb\ incurs some per-update cost but compensates for it with significantly better sample efficiency and final performance. Training time and sample efficiency reflect different practical costs: a method may run slower per step yet require far fewer episodes to succeed (as with \algoabb), which is valuable when interactions are expensive or data collection is slow (e.g., real robots, online multi-agent systems). Such added cost is also often acceptable in domains where miscoordination is expensive (e.g., drone swarms, multi-robot warehouses). Conversely, faster per-step but sample-inefficient methods may be suitable for inexpensive simulations where wall-clock time is critical (e.g., large-scale self-play).

In summary, our experiments provide a clear picture of \algoabb’s strengths and trade-offs. (1) It delivers state-of-the-art coordination performance, consistently outperforming strong graph-based methods and fully factorized value decomposition baselines across all MACO tasks. (2) Ablation results confirm the importance of the compositional design: layered refinement, parallel channels, and a diverse base relation bank are each critical for adaptively capturing inter-agent dependencies. (3) Sparse base graph initializations offer a practical way to reduce computation and can still yield satisfactory learning when designed with diversity; however, complete graphs as bases remain the most reliable and expressive default when no strong prior structure is available. (4) Capacity-scaled comparisons show that simply enlarging payoff networks or deepening standard GCNs in existing CG methods cannot match the expressive agent information integration achieved by MCG composition, confirming that \algoabb's gains stem from its design rather than model size alone. (5) While \algoabb\ adds per-step computational cost, it compensates by converging in much fewer episodes and reaching higher task quality. This can be a worthwhile trade-off when reliable multi-agent cooperation outweighs the need for less training time.

\section{Conclusion and Future Work} 
\label{sec:conclusion}


In this paper, we presented \textit{\algosmall} (\algoabb), a new approach for cooperative MARL that overcomes the limitations of value-decomposition and coordination graph methods in modeling complex, evolving inter-agent dependencies. \algoabb\ dynamically composes task-adaptive \emph{meta coordination graphs} to model potentially emerging agent interactions and guides information integration and factored Q-value learning for coordinated decision-making. Experiments across coordination-intensive benchmarks showed that \algoabb\ consistently outperforms many diverse MARL baselines. Further ablation and analytical studies confirm that its compositional design meaningfully improves coordination quality beyond simple model scaling, while our complexity analysis highlights the trade-off between coordination expressiveness and runtime efficiency. Future work can explore scaling to larger agent populations, mixed cooperative-competitive settings, and integrating off-policy learning or domain priors to further enhance efficiency and generalization.

\begin{acks}
This work is supported by DEVCOM Army Research Office
(ARO); Grant: W911NF2420194; and U.S. National Science Foundation (NSF); Grant:  OAC-2411446. Distribution Statement A: Approved for public release. Distribution is unlimited. 
\end{acks}

\bibliographystyle{ACM-Reference-Format}
\bibliography{aamas}

\cleardoublepage
\section*{Appendix}
\appendix 
\label{sec:appendix}


\noindent\rule{\linewidth}{1pt}

\noindent \textbf{Appendix A}: This section provides further details about evaluation domains used in the study. 

\noindent \textbf{Appendix B}: This section discusses further experiments of \algoabb's evaluation on the StarCraft Multi-Agent Challenge (SMACv2) and a discussion on generalization, strategy, and trade-offs. 

\noindent \textbf{Appendix C}: This section provides further details about algorithm implementations in the study. 

\noindent \textbf{Appendix D}: This section provides further details on some of the ablation study conducted in this paper. 

\noindent \textbf{Appendix E}: This section provides further details on the computational complexity of \algoabb; it compares training time, convergence, and model size across methods. 

\noindent\rule{\linewidth}{1pt} 

\section*{Appendix A: Task settings} 

\paragraph{Why MACO benchmark?} \cite{wang2022contextaware} It is designed to evaluate multi-agent reinforcement learning algorithms by presenting them with a series of complex and diverse coordination tasks. This benchmark draws from classic problems in the cooperative multi-agent learning literature, enhancing their difficulty. Each task represents a specific type of coordination challenge where agents must learn different coordination strategies. By increasing the complexity of these tasks, the MACO benchmark provides a rigorous framework for analyzing the performance and adaptability of multi-agent learning approaches in various cooperative scenarios. 

\paragraph{Gather} (Figure \ref{fig:envs}a) is an extension of the Climb Game \cite{wei2016lenient}. Gather increases the task's complexity by extending it temporally and adding stochasticity. Agents must navigate to one of three potential goal states: \( g_1 \), \( g_2 \), or \( g_3 \), corresponding to the actions \( a_0 \), \( a_1 \), and \( a_2 \) in Climb. At the start of each episode, one goal is randomly designated as optimal. Agents are randomly spawned, and only those near the optimal goal know its designation. The reward function is:\[R = 
\begin{cases} 
10 & \text{if all agents reach the optimal goal } g_1, \\
5 & \text{if all agents reach a non-optimal goal}, \\
-5 & \text{if only some agents reach the optimal goal}.
\end{cases}
\]

\paragraph{Disperse} (Figure \ref{fig:envs}b) has twelve agents who must select one of four hospitals to work at each timestep. The environment has a dynamic requirement where only one hospital needs a specific number of agents at any given time $t$. The task tests the agents' ability to distribute themselves efficiently according to the hospital's needs, with penalties for understaffing, i.e., when \(y_j^{t+1} < x_j^t\), where $y_j^{t+1}$ is the number of agents that went to hospital $j$ and $x_j^t$ is the required number of agents for that hospital. \[ R = min(y_j^{t+1} - x_j^t, 0) \]

\paragraph{Pursuit or ``Predator and Prey"} (Figure \ref{fig:envs}c) has ten predator agents that must capture randomly walking prey on a $10 \times 10$ grid. The environment is designed to test the agents' ability to coordinate their movements to successfully capture the prey, which requires simultaneous actions by at least two predators. The task is made more challenging by introducing penalties for failed capture attempts. \[ R = 
\begin{cases} 
1 & \text{if prey is captured by two agents} \\
-1 & \text{if only one agent attempts to capture prey}
\end{cases}
\]

\paragraph{Hallway} (Figure \ref{fig:envs}d) is a multi-chain Dec-POMDP \cite{oliehoek2016concise} which extends the original Hallway problem \cite{wang2019learning}. Agents must coordinate their movements through a hallway to reach a goal state simultaneously. Each agent can observe its own position and choose to move left, move right, or stay still. The environment tests the agents' ability to synchronize their actions in the face of limited observability and potential conflicts when multiple groups attempt to reach the goal simultaneously. \[ R = 
\begin{cases} 
1 & \text{if agents in the same group} \\ 
& \text{reach goal $g$ simultaneously,} \\
-0.5 \times n_g & \text{if $n_g > 1$ groups attempt to} \\
                & \text{move to $g$ at the same time,}
\end{cases}
\]
where $n_g$ is the number of groups attempting to reach the goal simultaneously.

\section*{Appendix B: Additional experiments: Scaling to the StarCraft Multi-Agent Challenge (SMACv2)} 

\begin{figure*}[!htp]
    \centering
    \includegraphics[width=0.95\linewidth]{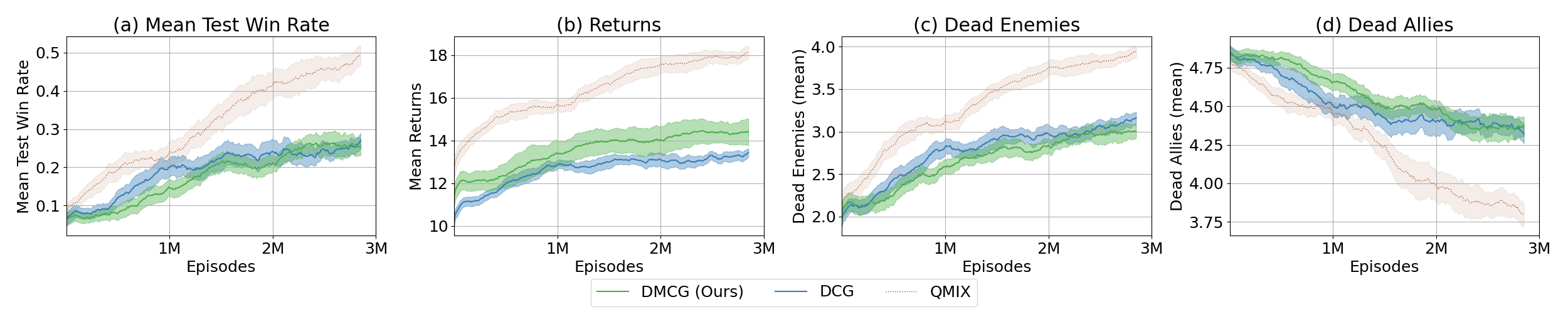} 
    \caption{Scaling \algoabb\ to SMACv2, highlighting key metrics such as test win rate, returns, and number of dead allies and enemies. \edits{The results suggest that \algoabb\ adopts a cautious and strategic approach, prioritizing survival and long-term returns over immediate aggression}.}
    \label{fig:res:smacv2}
\end{figure*}

\paragraph{StarCraft Multi-Agent Challenge (SMACv2)} (Figure \ref{fig:envs}e) is a challenging benchmark for MARL due to its combination of high-dimensional, partially observable environments, diverse unit types, and dynamic, stochastic elements. SMACv2 \cite{ellis2023smacv} introduces procedurally generated scenarios that require agents to generalize to unseen settings during evaluation, addressing the lack of stochasticity and partial observability in the original SMAC \cite{samvelyan2019starcraft}, which previously allowed near-perfect performance with simpler open-loop policies. 

\paragraph{\algoabb\ on SMACv2.} We now evaluate \algoabb’s scalability on SMACv2 using a 5v5 Protoss scenario featuring Stalkers, Zealots, and a Colossus. Agents spawn at random positions on a 32x32 map at difficulty level 7. \newedits{As shown in Figure~\ref{fig:res:smacv2}, \algoabb\ performs comparably to DCG, while QMIX outperforms both. This is consistent with prior observations that value-decomposition methods excel in SMACv2 despite lacking explicit coordination modeling \cite{yu2022surprising}. This is likely because SMACv2 scenarios do not exhibit strong miscoordination patterns such as relative overgeneralization, making it less suitable for benchmarking the advantages of structured coordination \cite{bohmer2020deep,gupta2021uneven}. We compare only with DCG here to isolate the impact of meta coordination graphs over traditional, static coordination graphs. Unlike DICG, which implicitly models dynamic interactions but lacks explicit graph structure, DCG provides a more direct baseline for evaluating structural improvements in \algoabb. The results confirm that \algoabb\ retains strong performance even in larger domains, highlighting the robustness of our method.} 

On further analysis of these results, we examined additional metrics, including test mean returns, average number of dead allies, and average number of dead enemies. We observed that \algoabb\ resulted in slightly fewer dead enemies on average compared to DCG, but it also maintained a lower number of dead allies. Moreover, \algoabb\ achieved a higher mean return over the episodes. We speculate that \algoabb\ agents are striking a better balance between aggression and preservation in this scenario. The slightly lower number of enemy casualties might suggest a more cautious or strategic approach, where agents prioritize survival and long-term success over immediate, aggressive tactics. This approach seems to result in fewer casualties among allies, which could contribute to the higher overall returns observed. \algoabb, through its ability to capture higher-order and indirect relationships among agents, is likely leading to a different cooperative strategy here, one which is trying to reduce casualties instead of focusing on winning. A more detailed study may be needed to further analyze \algoabb\ in this context, which we leave to future work.


\section*{Appendix C: Overall experimental setup} 

The code is available at {\color{blue}\url{https://github.com/Nikunj-Gupta/dmcg-marl}}. The repository contains all configuration files and hyperparameter settings used in our experiments. A zipped copy of the source code is also included with this appendix in the supplementary material. All baseline MARL algorithms (VDN, QMIX, CASEC, NLCG, GACG, VAST, DCG, DICG) were implemented using the PyMARL~\cite{rashid2018qmix} and PyMARL2~\cite{hu2023rethinking} frameworks, with necessary adaptations made for compatibility with our evaluation environments. DCG was implemented using the authors’ official repository at \url{https://github.com/wendelinboehmer/dcg}, while CASEC and NLCG were based on their respective official implementations. DICG was used in a centralized training–centralized execution (CTCE) setting with QMIX as the base policy learner, following its original training formulation. Each experiment was executed using four random seeds to ensure statistical robustness. Reported results show the mean (solid lines) and standard deviation (shaded regions) across these runs.

\section*{Appendix D: Ablation studies: Details on Coordination Graph Initialization Topologies} 

In our ablation studies, we evaluate the impact of different initial coordination graph structures on the performance of \algoabb. Each configuration initializes the Meta Coordination Graphs (MCGs) with multiple base graphs, each representing a distinct hypothetical interaction pattern among agents. Below, we detail the specific structures used:
\begin{itemize}
    \item \textbf{\algoabb (Full):} Initializes $K$ fully connected graphs, one for each hypothetical edge type. Every agent is connected to every other agent in each graph, allowing the model to start with maximal expressivity and prune unnecessary edges during learning. 
    \item \textbf{\algoabb-line:} Each base graph is a line topology where agents are connected sequentially (e.g., agent 1 to agent 2, agent 2 to agent 3, and so on). This encourages encoding of local and directional dependencies. 
    \item \textbf{\algoabb-star:} Each base graph has a central “hub” agent connected to all others, simulating hierarchical communication or leadership roles. We cycle through all agents as hubs across the multiple base graphs to avoid bias. 
    \item \textbf{\algoabb-cycle:} Agents are arranged in a closed loop in each base graph. This enforces uniform pairwise interactions with a cyclic structure, representing settings like patrol formations or ring-based topologies. 
    \item \textbf{\algoabb-kite:} Each base graph combines a tightly connected triangle (three agents) with a short “tail” (one or two agents connected to a single triangle node). This produces a semi-sparse structure mixing dense local interactions and limited peripheral links, modeling localized coordination with weakly connected outliers.
    \item \textbf{\algoabb-fan:} Each base graph consists of a “fan” structure, formed by connecting one central node to a chain of other agents (e.g., a star–line hybrid). This configuration promotes directional but shared communication, simulating cases where one agent disseminates information sequentially through a subset of others.
    \item \textbf{\algoabb-sparse:} A mixed initialization containing five distinct graphs: fully connected, line, star, cycle, and kite. The kite graph is a sparsely connected structure combining a triangle (3 agents) with a “tail” (1-2 additional agents connected to only one of the triangle nodes), encouraging diverse but localized interaction patterns. This configuration is designed to provide heterogeneity in inductive bias without overwhelming connectivity. 
\end{itemize}

We now investigate the influence of these different initial coordination graph topologies to validate the robustness and adaptability of \algoabb’s methodology, particularly given our standard use of fully-connected initialization. While our primary methodology leverages dense initial representations to facilitate extensive exploration of agent interactions, these ablations specifically assess how effectively \algoabb\ operates under various structured and sparse initial configurations. 


Figure~\ref{fig:res:gather_ablations} (in the main paper) demonstrates that \algoabb\ achieves superior performance when initialized with fully connected graphs, converging more rapidly and attaining a higher final win rate. This result underscores the importance of maximal initial connectivity, allowing the algorithm to comprehensively explore possible interactions early in training. Interestingly, the \algoabb-sparse\ configuration outperforms individual structured configurations (line, cycle, star), demonstrating that a diverse set of initial graphs facilitates richer composition and learning of emergent patterns. However, \algoabb-sparse\ still does not match the performance of the fully connected variant, indicating that while diverse structures offer benefits, the expressive power provided by dense initial connectivity remains superior. Notably, \algoabb-sparse still performs better than several state-of-the-art baselines, confirming the overall robustness of the methodology. 

These ablations thus validate \algoabb's robustness: although explicitly designed for maximal initial expressivity, it also effectively handles sparser initializations. This flexibility is especially valuable in practical scenarios, where domain-specific knowledge may suggest particular interaction structures. Nonetheless, our results advocate fully-connected initialization as a powerful default approach. 

\section*{Appendix E: Further discussion on computational complexity} 

\begin{table*}[h]
\centering
\caption{Comparison of training efficiency and model complexity across coordination graph-based MARL algorithms. Each entry reports the achieved metric, number of episodes to achieve peak performance (K), total wall-clock training time (hours), and number of agents used in that environment. Value-based methods are excluded due to poor performance.}
\label{tab:complexity_comparison}
\resizebox{\linewidth}{!}{%
\begin{tabular}{lcccccc}
\toprule
\textbf{Algorithm} & \textbf{Environment} & \textbf{Metric (Optimal Value)} & \textbf{Episodes (K)} & \textbf{Time (hrs)} & \textbf{\# Agents} \\
\midrule
\textbf{DMCG (Ours)} & Gather   & Win \% = 98\%    & 180  & 9   & 5  \\
\textbf{DICG}         & Gather   & Win \% = 97\%    & 800  & 2.5 & 5  \\
\textbf{DCG}          & Gather   & Win \% = 78\%    & 200  & 1.5 & 5  \\
\textbf{CASEC}        & Gather   & Win \% = 67\%    & 1000 & 5   & 5  \\
\textbf{NLCG}         & Gather   & Win \% = 99\%    & 1000 & 6   & 5  \\
\textbf{GACG}         & Gather   & Win \% = 83\%    & 650  & 3   & 5  \\
\textbf{VAST}         & Gather   & Win \% = 85\%    & 850  & 2   & 5  \\
\midrule
\textbf{DMCG (Ours)} & Disperse & Return = -1.4     & 100  & 15  & 12 \\
\textbf{DICG}         & Disperse & Return = -0.4     & 200  & 1   & 12 \\
\textbf{DCG}          & Disperse & Return = -2.0     & 150  & 5   & 12 \\
\textbf{CASEC}        & Disperse & Return = -1.8     & 175  & 1.5 & 12 \\
\textbf{NLCG}         & Disperse & Return = -2.2     & 50   & 4   & 12 \\
\textbf{GACG}         & Disperse & Return = -1.0     & 250  & 2   & 12 \\
\textbf{VAST}         & Disperse & Return = -3.0     & 150  & 1   & 12 \\
\midrule
\textbf{DMCG (Ours)} & Pursuit  & Prey Caught = 4.6  & 200  & 24  & 10 \\
\textbf{DICG}         & Pursuit  & Prey Caught = 0.0  & 300  & 8   & 10 \\
\textbf{DCG}          & Pursuit  & Prey Caught = 4.4  & 300  & 12  & 10 \\
\textbf{CASEC}        & Pursuit  & Prey Caught = 3.5  & 225  & 4   & 10 \\
\textbf{NLCG}         & Pursuit  & Prey Caught = 3.0  & 300  & 7   & 10 \\
\textbf{GACG}         & Pursuit  & Prey Caught = 0.0  & 300  & 3   & 10 \\
\textbf{VAST}         & Pursuit  & Prey Caught = 0.0  & 300  & 1   & 10 \\
\midrule
\textbf{DMCG (Ours)} & Hallway  & Win \% = 99\%      & 300  & 30  & 12 \\
\textbf{DICG}         & Hallway  & Win \% = 82\%      & 450  & 3   & 12 \\
\textbf{DCG}          & Hallway  & Win \% = 98\%      & 325  & 14  & 12 \\
\textbf{CASEC}        & Hallway  & Win \% = 5\%       & 450  & 5   & 12 \\
\textbf{NLCG}         & Hallway  & Win \% = 65\%      & 450  & 5   & 12 \\
\textbf{GACG}         & Hallway  & Win \% = 2\%       & 450  & 3   & 12 \\
\textbf{VAST}         & Hallway  & Win \% = 0\%       & 450  & 2   & 12 \\
\bottomrule
\end{tabular}
}
\end{table*}

Table~\ref{tab:complexity_comparison} provides a detailed comparison of training efficiency and computational characteristics across major coordination graph-based MARL methods, including \algoabb. For each environment, it reports the performance metric (e.g., win rate, return, prey caught), the optimal value achieved, the number of training episodes and wall-clock time required to reach that value, the number of agents involved, and the model’s parameter count. This structured analysis enables a clear examination of the trade-offs between computational cost and coordination performance across different algorithms. Although \algoabb\ occasionally incurs higher training time in certain environments, it consistently achieves the highest or near-highest optimal values, often requiring fewer episodes to converge, demonstrating both stability and strong sample efficiency. For instance, in the \textit{Gather} and \textit{Pursuit} environments, DMCG reaches superior coordination performance significantly faster than DCG, DICG, and CASEC, despite its more expressive architecture.

While \algoabb\ introduces moderate additional computation due to its multi-channel and compositional reasoning layers, the corresponding performance improvements justify the overhead, especially in tasks where miscoordination is costly. In real-world applications such as autonomous drone swarms for disaster response or multi-robot warehouse systems, slight coordination errors can lead to coverage gaps or inefficiencies. In such scenarios, the marginal increase in computation is a worthwhile trade-off for the substantial gains in cooperative behavior and decision quality offered by \algoabb.

This comparison focuses on structurally comparable coordination graph-based algorithms (DCG, DICG, CASEC, NLCG, GACG, and VAST) since they share the same underlying value factorization framework as \algoabb. Value-based decomposition methods, such as QMIX and VDN, are excluded because they consistently fail to achieve competitive coordination performance in these environments. Their limited representational capacity prevents them from capturing higher-order dependencies between agents, leading to persistent miscoordination and suboptimal joint actions. Consequently, reporting their 'optimal values' would not yield meaningful insights in this training efficiency comparison. Nevertheless, we include these value-based baselines in our main performance plots to illustrate their limitations and provide a comprehensive empirical landscape.

\end{document}